\documentclass[letterpaper, 10 pt, conference]{ieeeconf}  %

\IEEEoverridecommandlockouts                              %

\usepackage{graphics} %
\usepackage{epsfig} %
\usepackage{mathptmx} %
\usepackage{times} %
\usepackage{graphicx}
\usepackage{booktabs}
\usepackage{tikz}
\usepackage{comment}
\usepackage{amsmath,amssymb} %
\usepackage{color}
\usepackage{xspace}
\usepackage{booktabs}
\usepackage{array,multirow}
\usepackage{algpseudocode}
\usepackage{algorithm}
\usepackage{caption}
\usepackage{hyperref}
\usepackage{subcaption}

\DeclareMathOperator*{\argmax}{argmax}

\input{macros}

\title{\LARGE \bf
Lidar Panoptic Segmentation and Tracking without Bells and Whistles
}

\author{Abhinav Agarwalla$^{1}$, Xuhua Huang$^{1,2}$, Jason Ziglar$^{3}$,
        Francesco Ferroni$^{3}$,\\ Laura Leal-Taixé$^{4}$, James Hays$^{5}$, Aljoša Ošep$^{1}$ and Deva Ramanan$^{1}$
       \\
       $^1$\small Carnegie Mellon University\quad 
       $^2$\small Meta AI\quad
       $^3$\small Argo AI\quad
       $^4$\small TU Munich\quad
       $^5$\small Georgia Institute of Technology\\
       }

\begin{document}

\maketitle
\thispagestyle{empty}
\pagestyle{empty}

\begin{abstract}
State-of-the-art lidar panoptic segmentation (LPS) methods follow ``bottom-up" segmentation-centric fashion wherein they build upon semantic segmentation networks by utilizing clustering to obtain object instances.
In this paper, we re-think this approach and propose a surprisingly simple yet effective detection-centric network for both LPS and tracking. Our network is modular by design and optimized for all aspects of both the panoptic segmentation and tracking task.
One of the core components of our network is the object instance detection branch, which we train using point-level (modal) annotations, as available in segmentation-centric datasets. 
In the absence of amodal (cuboid) annotations, we regress modal centroids and object extent using trajectory-level supervision that provides information about object size, which cannot be inferred from single scans due to occlusions and the sparse nature of the lidar data. We obtain fine-grained instance segments by learning to associate lidar points with detected centroids. We evaluate our method on several 3D/4D LPS benchmarks and observe that our model establishes a new state-of-the-art among open-sourced models, outperforming recent query-based models.

\end{abstract}

\section{Introduction}

Lidar panoptic segmentation (LPS) is the task of labeling all 3D points with distinct semantic classes and instance IDs. 
This is directly relevant to online streaming robot operation, as robots need to be aware of both scene semantics and surrounding dynamic objects in order to navigate safely. 

\begin{figure}[t]
\centering
\includegraphics[width=1.0\linewidth]{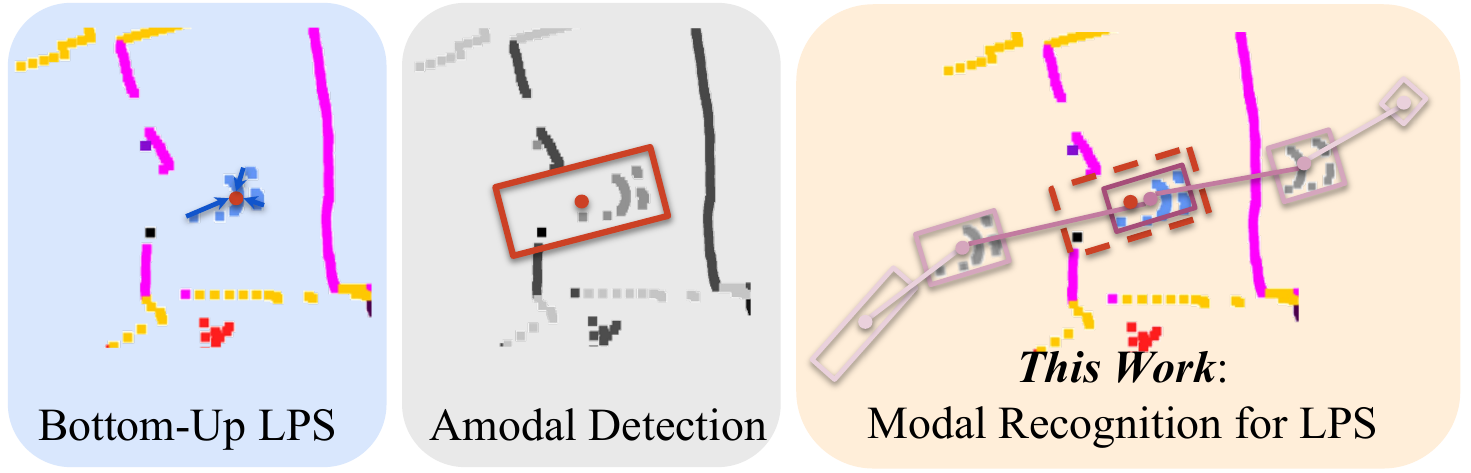}
\caption{
State-of-the-art LPS methods (\textit{left}) learn to group points in a bottom-up fashion, while state-of-the-art 3D object detectors (\textit{center}) detect objects as \textit{amodal} centers in the bird's-eye representation of the scene, followed by \textit{amodal} 3D bounding box regression. 
In this paper, we re-purpose the \textit{latter} for the \textit{former}. Our method in parallel classifies points (\textit{semantic segmentation}), detect \textit{modal} instance centers (\textit{modal instance recognition}) and their velocities (\textit{modal instance tracking}). %
}
\label{fig:teaser}
\end{figure}

While state-of-the-art 3D detection and tracking methods detect objects in top-down fashion (Fig.~\ref{fig:teaser}, \textit{center}) and regress full object extent and orientation/velocity~\cite{yin2021center,yan2018second,Lang19CVPR,Shi19CVPR}, lidar instance and panoptic segmentation (Fig.~\ref{fig:teaser}, \textit{left}) follow ``bottom-up" segmentation-centric philosophy~\cite{aygun21cvpr,gasperini2020panoster,hong2021lidar,razani2021gp,li2022panoptic}, that does not require reasoning about the full extent of 3D bounding boxes. Instead, these \textit{segmentation-centric} first perform per-point semantic classification and then learn to group points corresponding to \textit{thing} classes into instances in a bottom-up fashion. 

In this paper, we question the established narrative that bottom-up grouping is the only design pattern for LPS and propose \methodabr (\method), a surprisingly simple yet effective \textit{detection-centric} approach for \textit{lidar panoptic segmentation}~\cite{Behley21icra} and tracking~\cite{aygun21cvpr}. 
Concretely, we base our method on CenterPoint~\cite{yin2021center}, designed for top-down lidar-based 3D object detection. As in CenterPoint, we first encode a point cloud sequence using a sparse 3D convolutional backbone~\cite{yan2018second,zhou2018voxelnet,yin2021center,zhu2020cylindrical} and flatten the bottleneck layer into a bird's eye view (BEV) representation of the point cloud that we use to detect objects as points. 
To obtain dense, per-point semantic interpretation and instance interpretation of point clouds, we add a 3D decoder head to our network that un-projects this representation and up-samples it back to the original resolution to perform per-voxel semantic classification. 
While 3D bounding boxes (\cf,~\cite{yin2021center}) can directly be regressed from per-point bottleneck features, necessary fine-grained details needed for point-wise classification and instance segmentation are lost. Therefore, inspired by the instance segmentation branch of modern two-stage instance segmentation methods, we add a second-stage instance segmentation network that determines the \textit{membership} of points to their respective instance centers provided by the instance recognition branch. Finally, we obtain spatio-temporal instance labels by additionally learning to regress modal offset vectors used for scan-to-scan instance association~\cite{yin2021center}. 

The \textbf{motivation} for this design is two-fold. Firstly, LPS methods should maximize all aspects of the panoptic segmentation and tracking task, \ie, (i) object recognition, (ii) instance segmentation, (iii) per-point semantic classification, and (iv) tracking, all explicitly captured via different modules in our network, supervised with corresponding loss functions. 
Second, this network is modular by design -- while object detection, point classification, instance segmentation, and velocity regression components all share a common feature extractor, different components are disentangled. This, in principle, allows us to investigate the performance of each module separately, which is important for model interpretability that is crucial in robotics applications. Importantly, such a model can be  trained in the future on multiple datasets with different levels of supervision (as densely-labeled data is expensive to obtain), or replace different modules with stronger counterparts to boost the performance further.

Importantly, 3D detectors, such as CenterPoint, used as a base for our model, rely on \emph{amodal} 3D bounding box supervision that enclose the full extent of the object, not only the visible portion. Such labels are not necessarily available in segmentation-centric semantic/panoptic segmentation datasets~\cite{Behley19ICCV,Behley21icra}.\footnote{With exception of nuScenes, which also includes \textit{amodal} 3D object detection labels.} 
To remedy this, we show we can leverage track-level (temporal) information to reason about the full extent of objects during the network training, and, as our experiments confirm, we alleviate the need for \emph{amodal} labels. 
From this perspective, our method \methodabr \textit{marries} object instance recognition and semantic segmentation in a single modular network suitable for 3D and 4D lidar panoptic segmentation and can be trained solely from temporal point-level (modal) supervision. This makes our method versatile enough for a thorough evaluation on multiple benchmarks for 3D/4D LPS on Panoptic nuScenes~\cite{fong21arxiv} and SemanticKITTI~\cite{Behley19ICCV,Behley21icra} datasets. 

In \textbf{summary}, (i) we propose a 3D/4D lidar segmentation network that unifies per-point semantic segmentation with modal object recognition and tracking in a single network.
We (ii) detect instances via \textit{modal} point-based temporal supervision and segment them with our novel binary instance segmentation network that determines point-to-detection membership based on BEV and per-point semantic features. Finally, we (iii) show the effectiveness of our method on various benchmarks for 3D/4D LPS. This confirms that our top-down approach based on modal recognition is highly effective for both 3D and 4D lidar panoptic segmentation and may directly impact design patterns used in future developments in this field of research. Our code, along with experimental data, is available at \href{https://mostlps.github.io}{https://mostlps.github.io}.

\section{Related Work}
In this section, we summarize relevant work in 3D object detection, tracking, and semantic and panoptic segmentation for lidar point clouds.

\PAR{Semantic segmentation.}
Advances in deep representation learning on unordered point sets~\cite{Qi17CVPR_pointnet} enable direct encoding of raw, unstructured point clouds to estimate fine-grained per-point semantic labels~\cite{Qi17CVPR_pointnet,Qi17NIPS,Thomas19ICCV,hu20cvpr}. Alternatively, several methods~\cite{Wu18ICRA,Wu19ICRA,Milioto19IROS,aksoy2020salsanet,razani2021lite,li2021multi} operate on a spherical projection of point cloud (\ie, range images), and provide an excellent trade-off between accuracy and efficiency, important in robotics scenarios. 
State-of-the-art methods rely on voxel grids in conjunction with sparse convolutions~\cite{choy20194d,tang2020spvnas}. 
To efficiently encode sparse lidar point clouds, Cylinder3D~\cite{zhu2020cylindrical} performs a cylindrical partition and proposes asymmetrical 3D convolution networks, followed by point refinement. We similarly adopt a sparse voxel grid-based backbone for point-based classification, a sub-task of panoptic segmentation. 

\PAR{Panoptic segmentation.} 
Seminal methods for lidar panoptic segmentation follow a top-down approach inspired by the early image-based baselines~\cite{Kirillov19CVPR}. These approaches train separate networks for semantic segmentation and object detection, followed by heuristic result fusion~\cite{Behley21icra}.
However, recent trends show that in the lidar domain, bottom-up, methods~\cite{jiang20cvpr_pointgroup,han2020occuseg,Engelmann20CVPR,milioto2020iros,aygun21cvpr,gasperini2020panoster,hong2021lidar,razani2021gp,zhou2021panoptic}, including recent query-based networks~\cite{marcuzzi2023ral}, obtain state-of-the-art results. Are bottom-up methods based on point grouping and cross-attention de-facto go-to approaches for lidar panoptic segmentation? We suggest this is not necessarily the case.

\PAR{4D lidar panoptic segmentation.} Recently introduced 4D lidar panoptic segmentation~\cite{aygun21cvpr,fong21arxiv} extends lidar panoptic segmentation to the temporal domain, which requires sequence-level understanding.
4D-PLS~\cite{aygun21cvpr} poses this task as bottom-up spatio-temporal point grouping, while MOPT~\cite{hurtado2020mopt} and CA-Net~\cite{marcuzzi2022contrastive} segment instances in individual scans and associate them across time. Our proposed method is flexible and can generalize to utilize either single-scan or a multi-scan lidar sweep for both 3D and 4D lidar panoptic segmentation in one single unified network.

\PAR{(A)modal object localization.} Amodal bounding boxes (in 2D or 3D) encapsulate the full extent of the object, regardless of whether the full object is visible or not. This approach has origins in object detection~\cite{Everingham10IJCV} and requires the hallucination of bounding boxes during the annotation process. Recent works on 3D object detection~(\cite{yin2021center, Lang19CVPR, zhou2018voxelnet,yan2018second,liu2021iccv}) specifically utilizing amodal bounding boxes. Boxes can be hallucinated by annotators~\cite{Everingham10IJCV}, obtained via linear interpolation in sequences~\cite{dendorfer20ijcv} or in 3D using SLAM/structure from motion~\cite{Geiger12CVPR}. Alternatively, the recognition task can be posed as localization of the visible portion of the object (modal recognition), common in segmentation-centric tasks~\cite{Lin14ECCV,Voigtlaender19CVPR,Behley19ICCV}. Modal annotations do not require hallucination of unobserved regions and are thus less sensitive to localization errors and less expensive in terms of annotation costs. In this paper, we demonstrate using modal annotations can achieve competitive performance compared to existing works built on amodal annotations.

\section{MOST: Modal Segmentaton and Tracking}

\label{sec:method}

\begin{figure*}[ht]
\centering
\includegraphics[width=0.8\linewidth]{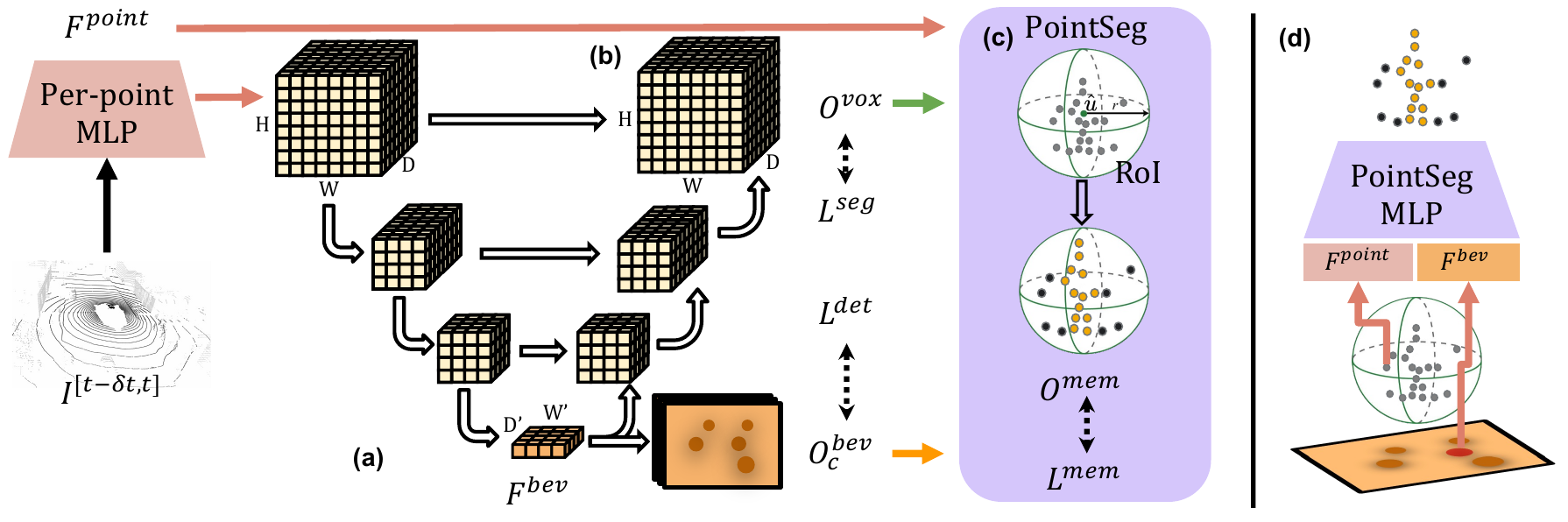}
\caption{\methodabr overview. We accumulate a point cloud sequence and encode it using a voxel grid-based encode-decoder backbone (\cf, \cite{zhou2018voxelnet,zhu2020cylindrical}). After accumulating encoded points in a 3D voxel grid, (\textit{a}) we down-sample the volume (via sparse 3D convolutions and pooling layers), and flatten the bottleneck layer along height-axis to obtain BEV representation (followed by 2D convolutional layers, similar to~\cite{yin2021center}). We use this representation to detect objects as (modal) points and regress (modal) offsets for temporal association. Our decoder (\textit{b}) consists of several up-sampling layers to obtain fine-frained, voxel-level semantic predictions. Our instance segmentation network, \memfunc (\textit{c}) performs binary classification within regions of interest (RoI) centered around predicted centers to obtain object instances. (\textit{d}) \memfunc utilizes point and center features as input to produce panoptic segmentation results.}
\label{fig:method}
\end{figure*}

In this section, we present \methodabr (\method) for lidar panoptic segmentation of point clouds and point cloud sequences. Lidar panoptic segmentation methods must predict semantic class and a unique instance identity label for each point in a point cloud (sequence). This task is especially challenging in the temporal domain because objects may become occluded or may exit or (re)-enter the sensing area. 
We first cover an overview of \methodabr, followed by a discussion of all key components. 

\PAR{Overview.}
A visual overview of \methodabr is presented in Fig.~\ref{fig:method}.
We base our network for lidar panoptic segmentation on an encoder-decoder-based U-net architecture~\cite{ronneberger2015u}. In particular, we build on a sparse voxel grid-based backbone~\cite{zhou2018voxelnet}, which encodes points with a shared multi-layer perceptron (MLP), and accumulates encoded points in voxels. We then apply several 3D sparse convolution layers to obtain a down-sampled BEV representation of the scene (Fig.~\ref{fig:method}\textit{a}), \ie, the bottleneck layer. 
Next, we detect object instances via the modal instance detection branch on top of the BEV representation. 
In parallel, our decoder upsamples the bottleneck layer back to the original voxel grid resolution via a series of 3D upsampling layers to obtain voxel-level semantic predictions~(Fig.~\ref{fig:method}\textit{b}). 
\textit{Finally}, our instance segmentation network, PointSegMLP (Fig.~\ref{fig:method}\textit{c}), determines which points belong to detected instances. To this end, PointSegMLP classifies points within instance-specific regions of interest (RoIs) centered around detected instances leading to panoptic segmentation predictions.

\PAR{Network architecture.}  The input to our network is a point cloud $\mathcal{I}^{t} = \{(x,y,z,intensity), \dots\} \in N \times \mathbb{R}^{4}$, where $N$ denotes the number of points. We accumulate input point clouds over a time window $[t - \delta t, t]$ to obtain \textit{4D point cloud} $\mathcal{I}^{[t-\delta t, t]} \in N' \times \mathbb{R}^{5}$, wherein the last dimension encodes the relative time $\delta t$. We encode points using a MLP to obtain per-point features $F^{point}$, which we accumulate in a regular 4D voxel grid $\mathbb{R}^{C\times H\times W\times D}$, where $H$, $W$, $D$ are dimensions of the bounding volume and  $C$ is the channel dimension. This 4D multichannel feature grid is processed with 3D convolutional encoders and decoders, following the sparse convolutional backbone of VoxelNet~\cite{zhou2018voxelnet}, which has proven successful for 3D object detection~\cite{yan2018second,zhou2018voxelnet,yin2021center} and semantic segmentation~\cite{zhu2020cylindrical}. 

For the \textit{modal detection} branch, we flatten voxel features along their height to obtain a BEV feature map $F^{bev} \in \mathbb{R}^{C' \times W' \times D'}$. We then apply 2D convolutional layers to reduce the channel dimension to $K$ output classes, followed by ReLU activation function to obtain \textit{modal} center heatmaps $O^{bev}_c \in \mathbb{R}^{K \times W' \times D'}$, followed by non-maxima suppression to obtain a set of detected instances.

To obtain point-precise \textit{semantic segmentation}, we up-sample the BEV feature map $F^{bev}$ via upsampling layers back to a voxel-grid representation to obtain voxel-level logits $O^{vox}$. In a U-net fashion, we add skip connections from downsampling layers to capture fine-grained features. We then classify voxels via the softmax classifier. 

Finally, to segment instances corresponding to predicted centers, we train an instance segmentation network (\memfunc) that predicts point-to-center \textit{memberships}. Given a predicted center $\hat{c} \in \mathbb{R}^3$, we compute a binary membership for all points $p \in RoI(\hat{c})$. 
To this end, we concatenate per-point features $F^{point}$ and predicted center BEV features $F^{bev}$ for each point-center pair. Next, we use concatenated features to determine instance memberships $O^{mem}(p, \hat{c}) \in [0,1]$ for all pairs. We detail all components of our network in the following paragraphs. 

\PAR{Semantic segmentation.} For voxel-level supervision, we obtain the supervisory signal from the current sweep $\mathcal{I}^{t}$ via majority voting to obtain %
$Y^{vox} \in \mathcal{K}^{H\times W\times D}$, where $\mathcal{K}$ denotes the set of all classes. Next, we apply per-voxel cross-entropy (CE) loss on-top of the voxel logits $O^{vox}$:
\begin{equation}
    L_{seg} = CE(Y^{vox}, O^{vox}).
\end{equation}
We note that even though we accumulate raw point clouds as $\mathcal{I}^{[t-\delta t, t]}$ as input to the encoder, the loss is only applied to voxels corresponding to $\mathcal{I}^t$. This is done by simply masking out loss corresponding to voxels in $\mathcal{I}^{[t-\delta t, t]}$ that do not belong to the current sweep $\mathcal{I}^t$. For point-level supervision, we utilize a point-refinement network~\cite{zhu2020cylindrical}. 
We obtain voxel features $O^{vox}$ at the point level, point features $F^{point}$, and BEV features $O^{bev}$ and train a linear layer using cross-entropy loss. During inference, we assign point-level predictions $O^{point}$ to all points within the voxel. 

\PAR{Modal object instance recognition.} 
Assuming access to only per-point semantic class and instance IDs for objects, we represent objects via statistics computed from observed points. More precisely, for a visible set of points $\mathcal{P}$ representing an instance, we define \textit{modal} center $c \in \mathbb{R}^3$ as the mean of $\mathcal{P}$, and \textit{modal} extent $r \in \mathbb{R}^3$ as the maximum distance of a point $p$ from $c$. 
Intuitively, modal extent encodes the visible extent of an object. 
Following~\cite{yin2021center}, we obtain BEV supervisory signal by constructing $K$ class-wise center heatmaps $Y^{bev}_c \in \mathbb{R}^{W' \times D'}$. We project modal centers and modal extents from 3D into the 2D BEV plane. Then, we place a 2D Gaussian centered at the projected modal centers with the projected radius as the variance. Since the projection collapses the center height information, we set an additional regression target $Y^{bev}_h \in \mathbb{R}^{W' \times D'}$ to localize the object in 3D. We then apply focal loss~\cite{lin2017focal} for \textit{thing} classes as in CenterPoint~\cite{yin2021center} for modal centers heatmaps~$O^{bev}_c$. In addition we apply an L1 regression loss on height~$O^{bev}_h$: 
\begin{equation}
    L_{det} = FocalLoss(Y^{bev}_{c}, O^{bev}_{c}) + |Y^{bev}_{h} - O^{bev}_{h}|. 
\end{equation}

\PAR{Estimating modal extent $r$.} We first compute the extent for instances at each time step via shrink-wrapping (SW). To this end, we estimate tight axis-aligned bounding boxes that enclose observed points. We compute SW axis-aligned box at time $t$ as: $r_t = \max\left\{|p-c|, p\in\mathcal{P}\right\}$, where $\mathcal{P}$ represents the set of observed points for this instance. 

Intuitively, humans reason about the object's extent by fusing information from multiple viewpoints. A sensor mounted on an autonomous vehicle similarly observes objects from different viewpoints over time. Therefore, we can derive more accurate object extent estimates $r$ by reasoning about object size over time. We utilize unique instance IDs, available in 4D panoptic segmentation datasets~\cite{Behley21icra,fong21arxiv} to obtain refined object extent estimates through \textit{temporal supervision}. To this end, we simply compute the maxima of all per-frame extent estimates for an object across time (MAX) to obtain a more precise estimate compared to naive SW: $r = \max\left\{r_t\right\}_{t=0}^{T}$. These are then used as target extents during the modal instance recognition branch training. This approach is especially beneficial for instances that contain only a few points—for example, cases where only a vehicle's front or rear bumper is visible. 

\PAR{PointSegMLP for instance segmentation.}
The modal recognition branch provides object center predictions, while the semantic decoder independently provides point-wise semantic predictions. 
The next step is to obtain \textit{instance-level} segmentation, \ie, modal point-precise segmentation masks for detected instances. We tackle instance segmentation by training a \textit{point membership} function, \memfunc, that determines for each point in the scene \textit{which} points correspond to \textit{which} detected centroids. 
This is analogous to image-based two-stage instance segmentation networks~\cite{He17ICCV}, that segment instances via binary classification for each anchor box. 

Consider a detected object instance $D$ with center $\hat{\mu}$ and class $\hat{k}$. We take all points $\mathcal{P} = \{p \in RoI(\hat{\mu},\hat{k})\}$ as \textit{in-range} points, wherein each $RoI$ is constructed from the predicted modal extents of the detected object. Next, we obtain features for the detected center and for all in-range points $p$~(see Fig.~\ref{fig:method}\textit{d}). The center features are comprised of its 3D position $\hat{\mu}$, semantic class $\hat{k}$ and BEV features $F^{bev}_{\hat{\mu}}$. The BEV features $F^{bev}_{\hat{\mu}}$ are obtained by first projecting $\hat{\mu}$ onto the BEV plane, followed by linear interpolation of $F^{bev}$ features at the projected point. Similarly, we compute point features using its 3D position $p \in \mathcal{P}$, predicted semantic label $O^{point}_p$ and the BEV features $F^{bev}_{p}$ for point $p$. In addition, we append $F^{point}_{p}$ features obtained from the backbone. The obtained center features are concatenated with the point features to obtain a feature representation for a point-center pair. Our per-point \memfunc, shared across all points, utilizes the obtained point-center features to determine per-point-to-center instance membership $O^{mem} \in [0,1]$. The architecture boils down to a light-weight MLP comprised of fully connected layers with batch normalization and ReLU activation. For supervision, we construct a binary ground truth membership $Y^{mem}$, points belonging to an instance $D$ are assigned $1$, and others $0$. We train \memfunc using binary cross-entropy loss (BCE): 

\begin{equation}
    L_{mem} = BCE(Y^{mem}, O^{mem}).
\end{equation}

\PAR{Modal panoptic tracking.} 
We follow CenterPoint~\cite{yin2021center} and concatenate point clouds before encoding them. Such spatio-temporal representations can be used to regress offset vectors $v \in \mathbb{R}^2$ and to obtain sequence-level lidar panoptic segmentation through greedy association. The difference, however, is that we estimate offset vectors using modal labels only. We construct ground truth velocity offsets $Y^{bev}_v$ for input point cloud $\mathcal{I}^t$ using $\mathcal{I}^{t-\delta t}$ and $\mathcal{I}^{t+\delta t}$. For each object, ground truth velocity offsets are computed through a centered difference between modal centers \ie, $(\mu^{t+\delta t} - \mu^{t-\delta t})/(2\delta t)$. %
The velocity offset predictions $O^{bev}_v$ when combined with per-point 3D panoptic segmentation leads to a unified, single-network top-down approach to 4D lidar panoptic segmentation. We train the velocity offset regression head using L1 loss:
\begin{equation}
    L_{track} = |Y^{bev}_{v} - O^{bev}_{v}|.
\end{equation}
 
\PAR{Putting everything together.} 
We train our network by minimizing the overall training objective, that is composed of modal detection loss $L_{det}$, semantic segmentation loss $L_{seg}$, instance segmentation loss $L_{mem}$, and optionally for sequences, modal velocity regression loss $L_{track}$:  %
\begin{equation}
    L_{total} = L_{det} + L_{seg} + L_{mem} + L_{track}.
\end{equation}

\PAR{Inference.}  During inference, we fuse segmentation branch predictions $O^{vox}$, modal center heatmaps $O^{bev}$, and point-center memberships $O^{mem}$ to obtain 4D panoptic predictions. We utilize segmentation labels predicted by the segmentation branch, and instance labels predicted by the modal centroid membership branch. We provide pseudocode in the supplementary, but summarize it here: using the segmentation branch predictions $O^{vox}$, we assign point-level predictions $O^{point}$ to all points within the voxel. Similarly, we apply non-maximum supression (NMS) over the predicted center heatmaps $O^{bev}$ to generate predicted modal centers $\hat{\mu}$. We then compute the membership of each point $p$ within the RoI of each center using \memfunc, resolving overlapping RoIs by the most confident center $\hat{u}^* = \argmax_{\hat{u}} (O^{mem}(p, \hat{u}))$. 
Next, we assign the predicted center label to all points that are its members, and assign a unique instance id. For all \textit{stuff} points, we directly utilize the predicted semantic label.
To extend our method to panoptic tracking, we associate instances across sweeps by using predicted center velocities $O^{bev}_v$, following the approach in~\cite{yin2021center}: we greedily form \textit{tracklets} by matching previous-sweep centers to current sweep centers with subtracted velocity offsets. 
Finally, all the instances of an object belonging to a \textit{tracklet} are assigned a temporally consistent unique ID.

\PAR{Implementation details.} We train our network in two stages. In the first stage, we optimize the modal detection and segmentation branch using $L_{det}$, $L_{seg}$, and $L_{track}$. Next, we freeze the first stage network and only train the second stage using a per-point $L_{mem}$ loss. The first stage network is trained with Adam optimizer with a learning rate of $1e^{-3}$, with a batch size of $8$. For the second stage of training of \memfunc, we utilize SGD optimizer with a learning rate as $5e^{-4}$. The network is trained for a total of $20$ epochs. Architecture: the per-point feature extraction layer and \memfunc are simple 4-layer MLPs with BatchNorm and ReLU layers. The voxel grid encoder downsamples the input point cloud by a factor of 8, while the decoder upsamples the bottleneck layer back to the original resolution. The modal detection branch comprises of two 3x3 convolution layers with ReLU activation layers. We accumulate previous $10$ frames. We do not employ any test time augmentation while reporting our results. Please refer to the supplementary for additional details. 

\begin{figure*}[t]
\centering
\setlength{\fboxsep}{0.3pt}%
\begin{subfigure}[b]{0.24\linewidth}
\centering
\rotatebox[origin=c]{90}{\fbox{\includegraphics[trim={4.5cm 0 0 1.5cm},clip,width=0.6\linewidth]{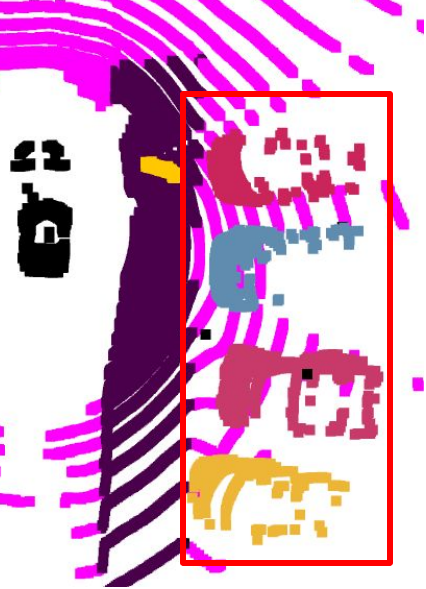}}} \\
    \vspace{-15pt}
    \caption{\footnotesize GT}
    \label{subfig:gt}
\end{subfigure}
\begin{subfigure}[b]{0.24\linewidth}
\centering
\rotatebox[origin=c]{90}{\fbox{\includegraphics[trim={4.5cm 0 0 1.5cm},clip,width=0.6\linewidth]{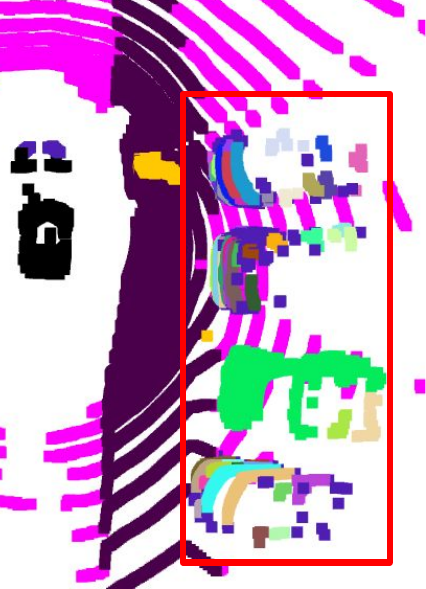}}} \\
    \vspace{-15pt}
    \caption{\footnotesize DS-Net}
    \label{subfig:dsnet}
\end{subfigure}
\begin{subfigure}[b]{0.24\linewidth}
\centering
\rotatebox[origin=c]{90}{\fbox{\includegraphics[trim={4.5cm 0 0 1.5cm},clip,width=0.6\linewidth]{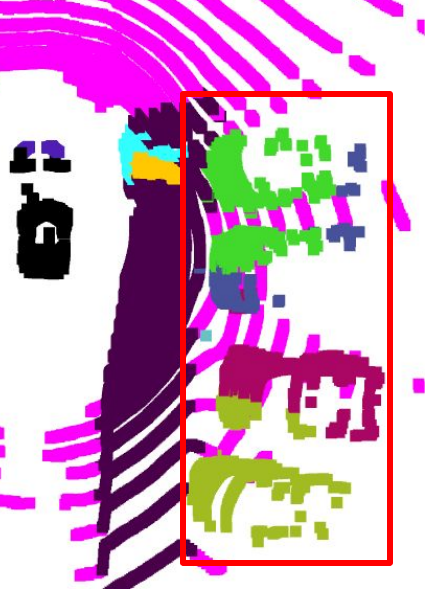}}} \\
    \vspace{-15pt}
    \caption{\footnotesize NN-Baseline}
    \label{subfig:nn}
\end{subfigure}
\begin{subfigure}[b]{0.24\linewidth}
\centering
\rotatebox[origin=c]{90}{\fbox{\includegraphics[trim={4.5cm 0 0 1.5cm},clip,width=0.6\linewidth]{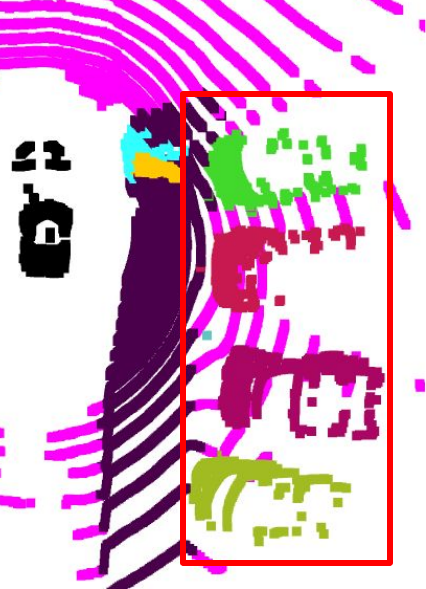}}} \\
    \vspace{-15pt}
    \caption{\footnotesize \methodabr}
    \label{subfig:ours}
\end{subfigure}
\caption{\textbf{Qualitative results:} We qualitatively compare the performance as compared to GT labels (Fig.~\ref{subfig:gt}). DS-Net~(Fig.~\ref{subfig:dsnet}) struggles with correctly segmenting multiple cars, leading to over-segmentation. Our \textit{modal} detection centric approach with the nearest neighbor heuristic (NN-baseline,  Fig.~\ref{subfig:nn} performs significantly better; however, we cannot correctly assign/segment \textit{fuzzy} points. By contrast, our full method, \methodabr (Fig.~\ref{subfig:ours}), correctly detects \textit{and} segments cars. 
} 
\label{fig:vis}
\end{figure*}

\begin{table}[t]
    \centering
    \caption{\textbf{Different strategies for amodalization:} 
    We compare different methods using the same segmentation backbone. We start with bottom-up centroid regression, as done in DS-Net. Naively creating a shrink-wrapped (SW) modal cuboid and training a top-down object-centric modal detector already improves performance, with or without sharing weights between the detection and segmentation branches. We also explore other ways of generating modal cuboids, including using a class-wise mean (CWM) cuboid dimension or taking the max dimension value across time (MAX). We see MAX performing the best when only modal annotations are available, while closing the gap with amodal annotations. The top results are \textbf{bolded}, while the second best are \underline{underlined}.} 
    \setlength{\tabcolsep}{3pt}
    \resizebox{1.0\columnwidth}{!}{%
        \begin{tabular}{l|ccc|ccc}
        \toprule
             Method & PQ & PQ$^{Th}$ & PQ$^{St}$ & mIoU & mIoU$^{Th}$ & mIoU$^{St}$ \\
            \midrule
              Offset reg. (DS-Net) & 68.2 & 66.1 & 71.6 & 75.5 & 71.8 & 81.7 \\
             \midrule
              Modal Det. (SW), w/o sharing & 72.2  & 71.7 & 72.9 & 77.2 & 74.5 & 81.7\\
             Modal Det. (SW), w/ sharing & 73.8 & 74.4 & 73.1 & 79.7 & 78.7 & 81.3 \\
             \midrule
             Modal Det. (DSB) & 70.1 & 68.4 & 72.9 & 77.6 & 75.2 & 81.3 \\
             Modal Det. (CWM) & 74.2 & 74.6 & \underline{73.5} & \underline{79.7} & 78.7 & \underline{81.3} \\
             Modal Det. (MAX) & \underline{77.1} & \underline{79.3} & \textbf{73.6} & \textbf{80.3} & \textbf{79.4} & \textbf{81.7} \\            
             \midrule
             Amodal Det. & \textbf{78.1} & \textbf{82.7} & 72.9 & 79.2 & \underline{78.8} & 79.7 \\
        \bottomrule
        \end{tabular}
    }
    \label{tab:topdown}
\end{table}

\section{Experimental Evaluation}
\label{sec:experimental}

\begin{table}[t]
    \centering
    \footnotesize
    \caption{\textbf{Different membership functions: }%
    We compare our \textit{PointSegMLP} with a simple yet surprisingly effective nearest neighbor heuristic (\textit{NN-baseline}). As can be seen, our learning-based \textit{PointSegMLP} based on 3D positions and semantic predictions already significantly outperforms the heuristic. While instance bird's-eye-view (BEV) features in isolation do not benefit our model, they further improve the performance when combined with per-point semantic features.}
    \setlength{\tabcolsep}{5pt}
        \begin{tabular}{l|c|c|c}
        \toprule
         Method & PQ & mIoU & Membership Acc (\%)\\
            \midrule
             Nearest Neighbor (baseline) &  70.9 & 77.2 & 76.3 \\ 
            Ours (semantic + geometric) & 72.9  & 77.5 & \textbf{95.4} \\
            \quad + BEV feat. & 72.6 & 77.4 & 95.0 \\
            \quad + Point feat. & \textbf{73.8} & \textbf{79.7} & \textbf{95.4} \\
        \bottomrule
        \end{tabular}
    \label{tab:pointseg_abla}
\end{table}

In this section, we first summarize our evaluation test-bed, including datasets, benchmarks and evaluation metrics used to conduct our experiments (Sec.~\ref{sec:testbed}). 
We next ablate various stages and design decisions of \methodabr 's network architecture for joint lidar panoptic segmentation and tracking (Sec.~\ref{sec:ablation}). 
Finally, we outline and discuss official benchmark results obtained on singe-scan and multi-scan (4D) lidar panoptic segmentation (Sec.~\ref{sec:bench}).

\subsection{Evaluation Test-Bed} 
\label{sec:testbed}

\PAR{Datasets.} We evaluate our work using SemanticKITTI~\cite{Behley19ICCV,Behley21icra} and Panoptic nuScenes~\cite{fong21arxiv} datasets, that contain per-point temporally-consistent semantic and instance labels. SemanticKITTI~\cite{Behley19ICCV,Behley21icra} contains $1.5h$ of lidar scans, recorded using a 64-beam sensor, and labels for 28 semantic classes. Panoptic nuScenes~\cite{fong21arxiv} contains $1,000$ short scenes with a 32-beam sensor. It contains labels for 32 semantic classes, with a labeling frequency of $2$Hz. 
For both datasets, we follow the official splits for training/validation and evaluate our final models on the hidden test set.

\begin{table*}[ht]
    \centering
    \caption{\textbf{LPS on nuScenes panoptic:} \methodabr is $1^{st}$ nuScenes val set and $2^{nd}$ on the test set ($1^{st}$ among open-source models). The best results are \textbf{bolded}, while the second best are \underline{underlined}.
    }
    \footnotesize
    \setlength{\tabcolsep}{5pt}
    \begin{tabular}{cl|cccc|ccc|ccc|c}
        \toprule
        & Method & PQ & PQ\textsuperscript{\textdagger} & RQ & SQ & PQ\textsuperscript{Th} & RQ\textsuperscript{Th} & SQ\textsuperscript{Th} & PQ\textsuperscript{St} & RQ\textsuperscript{St} & SQ\textsuperscript{St} & mIoU\\
            \midrule
            \parbox[t]{2mm}{\multirow{7}{*}{\rotatebox[origin=c]{90}{Validation}}}
            & Panoptic-PHNet~\cite{li2022panoptic} & \underline{74.7} & - & \underline{84.2} & \underline{88.2} & \underline{74.0} & - & - & \textbf{75.9} & - & - & \underline{79.7}\\
            & DS-Net~\cite{hong2021lidar} & 51.2 & - & 59.0 & 86.1 & 38.4 & 43.8 & \underline{86.7} & 72.3 & \underline{84.2} & \underline{85.0} & 73.5 \\ 
            & GP-S3Net~\cite{razani2021gp} & 61.0 & \underline{67.5} & 72.0 & 84.1 & 56.0 & 65.2 & 85.3 & 66.0 & 78.7 & 82.9 & 75.8 \\
            & Efficient-LidarPanopticSegmentation~\cite{sirohi2021efficientlps}\quad\quad & 62.0 & 65.6 & 73.9 & 83.4 & 56.8 & 68.0 & 83.2 & 70.6 & 83.6 & 83.8 & 65.6 \\
            & PolarSeg-Panoptic~\cite{zhou2021panoptic} & 63.4 & 67.2 & 75.3 & 83.9 & 59.2 & \underline{70.3} & 84.1 & 70.4 & 83.5 & 83.6 & 66.9 \\
            & MaskPLS~\cite{marcuzzi2023ral} & 57.7 & 60.2 & 66.0 & 71.8 & 64.4 & 73.3 & 84.8 & 52.2 & 60.7 & 62.4 & 62.5 \\
            & {\bf Ours} & \textbf{77.1} & \textbf{79.9} & \textbf{86.5} & \textbf{88.6} & \textbf{79.3} & \textbf{87.5} & \textbf{90.3} & \underline{73.6} & \textbf{84.9} & \textbf{85.7} & \textbf{80.3} \\
            \midrule
            \midrule
            \parbox[t]{2mm}{\multirow{4}{*}{\rotatebox[origin=c]{90}{Test}}}
            & Panoptic-PHNet~\cite{li2022panoptic} & \textbf{80.1} & \textbf{82.8} & \textbf{87.6} & \textbf{91.1} & \textbf{82.1} & \textbf{88.1} & \textbf{93.0} & \textbf{76.6} & \textbf{86.6} & \textbf{87.9} & \underline{80.2}\\
            & Efficient-LidarPanopticSegmentation~\cite{sirohi2021efficientlps}\quad\quad  & 62.4 & 66.0 & 74.1 & 83.7 & 57.2 & 68.2 & 83.6 & 71.1 & 84.0 & 83.8 & 66.7  \\
            & PolarSeg-Panoptic~\cite{zhou2021panoptic} & 63.6 & 67.1 & 75.1 & 84.3 & 59.0 & 69.8 & 84.3 & 71.3 & 83.9 & 84.2 & 67.0 \\
            & {\bf Ours} & \underline{76.1} & \underline{79.5} & \underline{85.1} & \underline{88.9} & \underline{77.4} & \underline{85.5} & \underline{90.3} & \underline{73.9} & \underline{84.5} & \underline{86.7} & \textbf{80.4} \\
        \bottomrule
        \end{tabular}
    \label{tab:nuscenes_panoseg}
\end{table*}

\begin{table*}[ht]
    \centering
    \caption{\textbf{LPS on SemanticKITTI:} \methodabr is a close $2^{nd}$ runner ($1^{st}$ among open-source models). Our method builds on the same encoder-decoder backbone as DS-Net; however, it significantly improves the results in terms of \textit{PQ} and \textit{mIoU}, thanks to our modal recognition branch and instance segmentation network. 
    } 
    \footnotesize
    \setlength{\tabcolsep}{5pt}
        \begin{tabular}{cl|cccc|ccc|ccc|c}
    \toprule
        & Method & PQ & PQ\textsuperscript{\textdagger} & RQ & SQ & PQ\textsuperscript{Th} & RQ\textsuperscript{Th} & SQ\textsuperscript{Th} & PQ\textsuperscript{St} & RQ\textsuperscript{St} & SQ\textsuperscript{St} & mIoU\\
        \midrule
        \parbox[t]{2mm}{\multirow{5}{*}{\rotatebox[origin=c]{90}{Validation}}}
        & DS-Net~\cite{hong2021lidar} & 57.7 & 63.4 & 68.0 & 77.6 & 61.8 & 68.8 & 78.2 & 54.8 & 67.3 & 77.1 & 63.5 \\ 
        & PolarSeg-Panoptic~\cite{zhou2021panoptic} & 59.1 & 64.1 & 70.2 & 78.3 & 65.7 & 74.7 & \textbf{87.4} & 54.3 & 66.9 & 71.6 & 64.5\\
        & Efficient-LidarPanopticSegmentation~\cite{sirohi2021efficientlps}\quad\quad  & 59.2 & 65.1 & 69.8 & 75.0 & 58.0 & 68.2 & 78.0 & \textbf{60.9} & 71.0 & 72.8 & 64.9 \\
        & GP-S3Net~\cite{razani2021gp} & \textbf{63.3} & \textbf{71.5} & \textbf{75.9} & \textbf{81.4} & \textbf{70.2} & \textbf{80.1} & \underline{86.2} & 58.3 & \textbf{72.9} & \textbf{77.9} & \textbf{73.0} \\
        & {\bf Ours} & \underline{63.1} & \underline{70.8} & \underline{73.1} & 79.2 & \underline{68.7} & \underline{75.7} & \underline{86.7} & \underline{58.9} & \underline{71.2} & \underline{73.7} & \underline{69.7}\\
        \midrule
        \midrule
        \parbox[t]{2mm}{\multirow{8}{*}{\rotatebox[origin=c]{90}{Test}}}
        & Panoptic-PHNet~\cite{li2022panoptic} & \textbf{61.5} & \underline{67.9} & \textbf{72.1} & \textbf{84.8} & \underline{63.8} & \underline{70.4} & \textbf{90.7} & 59.9 & 73.3 & \underline{80.5} & 66.0\\
        & SCAN~\cite{xu2022sparse} & \textbf{61.5} & \underline{67.5} & \textbf{72.1} & \textbf{84.5} & \underline{61.4} & \underline{69.3} & \textbf{88.1} & 61.5 & 74.1 & \underline{81.8} & 67.7\\
        & PolarSeg-Panoptic~\cite{zhou2021panoptic} & 54.1 & 60.7 & 65.0 & 81.4 & 53.3 & 60.6 & 87.2 & 54.8 & 68.1 & 77.2 & 59.5\\
        & DS-Net~\cite{hong2021lidar} & 55.9 & 62.5 & 66.7 & 82.3 & 55.1 & 62.8 & 87.2 & 56.5 & 69.5 & 78.7 & 61.6\\ 
        & Efficient-LidarPanopticSegmentation~\cite{sirohi2021efficientlps} & 57.4 & 63.2 & 68.7 & 83.0 & 53.1 & 60.5 & 87.8 & \underline{60.5} & \underline{74.6} & 79.5 & 61.4 \\
        & MaskPLS~\cite{marcuzzi2023ral} & 58.2 & 63.3 & 68.6 & 83.9 & 55.7 & 61.7 & 89.2 & 60.0 & 73.7 & 80.0 & 62.5 \\
        & GP-S3Net~\cite{razani2021gp} &60.0 & \textbf{69.0} & \textbf{72.1} & 82.0 & \textbf{65.0} & \textbf{74.5} & 86.6 & 56.4 & 70.4 & 78.7 & \textbf{70.8}\\
        & \textbf{Ours} & \underline{61.0} & 66.8 & \underline{72.0} & \underline{84.4} & 58.1 & 66.0 & \underline{88.1} & \textbf{63.2} & \textbf{76.3} & \textbf{81.7} & \underline{66.1} \\
    \bottomrule
    \end{tabular}
    \label{tab:semkitti_panoseg}
\end{table*}

\PAR{Tasks and evaluation metrics.}
For \textit{single-scan (3D) lidar panoptic segmentation}, we report well-established panoptic quality \textbf{PQ} metric~\cite{Kirillov19CVPR}, a soft version of the F1-score that treats \textit{thing} and \textit{stuff} classes in a unified manner. 
Following the official evaluation procedure, we set the minimum number of points on an instance to be $15$ for nuScenes and $50$ for SemanticKITTI.
We additionally report mean intersection-over-union (\textbf{mIoU})~\cite{Everingham10IJCV} that evaluates per-point semantic segmentation. 
For \textit{multi-scan (4D) lidar panoptic segmentation}, we report (lidar) segmentation and tracking quality \textbf{LSTQ}~\cite{weber21arxiv,aygun21cvpr}, evaluated as the geometric mean between mIoU and point association quality (\textbf{AQ}). AQ evaluates whether a point was associated with the correct instance in space and time. For Panoptic nuScenes, we additionally report the recently introduced panoptic tracking (\textbf{PAT}) metric, which combines PQ and LSTQ.

\subsection{Ablations} 
\label{sec:ablation}

\PAR{Modal recognition or center-offset regression?} We first study the impact of our top-down modal recognition-based approach to panoptic segmentation and compare it to a bottom-up center-offset regression approach by DS-Net~\cite{hong2021lidar}, for which code is available. This method predicts \textit{center offsets} followed by mean-shift clustering to obtain object instances. 
For comparison, we take the identical semantic segmentation network (Cylinder3D~\cite{zhu2020cylindrical}), but instead of offset regression and mean-shift clustering, we train a \textit{separate} modal instance recognition network, followed by our proposed instance segmentation network. We refer to this variant as \textit{Modal Det. (SW) w/o sharing} in Tab.~\ref{tab:topdown}.
As the semantic segmentation networks are identical, this experiment highlights the effectiveness of our modal detection branch. In this setting, we regress tightly-fitting ``shrink-wrap (SW)'' bounding boxes derived from segmentation labels. With this simple approach, we improve by $+4$ PQ points. %
\PAR{Joint training.} Next, we train a single network for \textit{joint} semantic segmentation and modal instance recognition and segmentation (as explained in Sec.~\ref{sec:method}), \ie, \textit{Modal Det. (SW) w/ sharing} in Tab.~\ref{tab:topdown}. 
This yields \textit{PQ} score of $73.8$ ($+0.6$), confirming the benefits of joint training of segmentation and modal recognition networks.

\PAR{Amodal recognition?} Next, we evaluate the impact of \textit{amodal} training on our detection component. 
Entry \textit{Amodal Det.} (Tab.~\ref{tab:topdown}) refers to a variant that trains the detection branch using such \textit{amodal} labels on nuScenes~\cite{fong21arxiv} which provides both segmentation labels and amodal boxes. 
We obtain PQ of $78.1$, $+4.3$ points compared to our SW \textit{modal} baseline. This is not surprising: \textit{amodal} labels provide additional supervisory signals in the form of orientation and full extent for each instance. 
This suggests that \textit{amodal} labels contain additional useful information that comes with additional annotation cost. \textit{This begs the question, can we close the gap using \textit{only} segmentation-level labels?}

\PAR{Closing the gap.} While regressing the full extent of the object is beneficial, we do not have access to this information (\eg, in SemanticKITTI~\cite{Behley19ICCV}). 
Can we do better than na\"ive \textit{shrink-wrapping} baseline ($73.8$ PQ)? First, we observe this gap is due to the sensitivity of the modal recognition head to occlusions and decreasing sensor resolution, resulting in a large number of instances containing only a few points. 
An obvious remedy is to simply \text{drop} small boxes (DSB), \ie, to exclude small tightly-fitting bounding boxes from training. However, this approach yields $70.1$ PQ, likely due to the exclusion of a large portion of training data. A better strategy is to replace small boxes with \textit{class-wise mean} (CWM) box sizes, which yields $74.2$ PQ. 
Finally, we utilize the sequential information and compute tightly-fitting boxes throughout the instance trajectory. Taking max dimension over time for supervision (MAX) produces PQ of $77.1$, which is reasonably close to full \textit{amodal} supervision. 
This implies that using sequential point-level labels, we can get better estimates of object extent resulting in better performance.

\PAR{PointSegMLP.} Next, we justify design decisions behind our instance segmentation network, \memfunc, that predicts binary point-center memberships for each detected object instance. In addition to reporting standard $PQ$ and $mIoU$, we also evaluate \textit{membership accuracy} (\textit{mem. acc.}), which computes the percentage of points within a given RoI that have been assigned to the correct center.

We first evaluate the performance of a simple geometric baseline, denoted as \textit{NN-baseline} in Tab.~\ref{tab:pointseg_abla}. This method assigns each point to its nearest \textit{semantically-compatible} detected center within the RoI of the detected center. We conclude that this simple baseline works surprisingly well, obtaining a $PQ$ of $70.9$. However, clearly, there is space for improvement in terms of \textit{mem. acc.} ($76.3$). While most points can be unambiguously assigned to the nearest instance, a certain percentage of \textit{fuzzy} points could be assigned to two or more instances. This motivates the usage of data-driven \memfunc to perform segmentation. 

Next, we compare this baseline to our \memfunc that learns to segment points. In the first variant (denoted as \textit{semantic + geometric}), we only utilize the 3D point coordinates of points and instance centers, along with their semantic predictions. We observe that just using geometrical features significantly outperforms the \textit{NN-baseline} with an improvement of $+19.1$ in terms of $mem. acc$, and this translates into $+2$ improvement in $PQ$. Solely adding bird-eye-view (BEV) feature (\textit{wrt.} detected instance) does not improve the performance; however, adding both instance \textit{BEV} feature and fine-grained \textit{point} features leads to a $+0.9$ increase in $PQ$. We also visualize the results in Fig.~\ref{fig:vis}.

\begin{table}[t]
    \centering
    \caption{\textbf{4D Lidar Panoptic Segmentation Benchmarks}. Our method is $1^{st}$ on nuScenes and $3^{rd}$ on SemanticKITTI. The top results are \textbf{bolded}, while the second best are \underline{underlined}. 
    MOT: tracking-by-detection~\cite{Weng20iros}, SFP: scene flow based propagation~\cite{mittal20cvpr}, PP: PointPillars 3D detector~\cite{Lang19CVPR}.} 
    \setlength{\tabcolsep}{5pt}
    \resizebox{\linewidth}{!}{%
        \begin{tabular}{cl|ccccc|c}
        \toprule
        & Method & $LSTQ$ & $PAT$ & $S_{assoc}$ & $S_{cls}$ & $PTQ$ & $PQ$\\
            \midrule
            \parbox[t]{2mm}{\multirow{6}{*}{\rotatebox[origin=c]{90}{SemanticKITTI}}} & RangeNet++~\cite{Milioto19IROS} + PP + MOT & 35.5 & -& 24.1 & 52.4 & -& -\\
            & KPConv~\cite{Thomas19ICCV} + PP + SFP & 38.5 & -& 26.6 & 55.9 & -& -\\
            & 4D-PLS~\cite{aygun21cvpr} & 56.9 & -& 56.4 & 57.4 & -& -\\%& - & - \\
            & Contrastive Association~\cite{marcuzzi2022contrastive} & \underline{63.1} & -& \underline{65.7} & \underline{60.6} & - & -\\%& - & - \\
            & 4D-StOP~\cite{kreuzberg20224d} & \textbf{63.9} & -& \textbf{69.5} & 58.8 & -& -\\
            & \textbf{Ours}  & 60.3 &-& 57.8 & \textbf{62.8} & -& -\\             
            \midrule
            \midrule
            \parbox[t]{2mm}{\multirow{5}{*}{\rotatebox[origin=c]{90}{nuScenes}}}
            & PanopticTrackNet~\cite{hurtado2020mopt} & 44.8 & 45.7 & 36.7 & 58.9 & 51.6 & 51.7 \\ 
            & 4D-PLS~\cite{aygun21cvpr} & 57.8 & 60.5 & 53.6 & 62.3 & 55.6 & 56.6\\
            & E-LPS~\cite{sirohi2021efficientlps} + Kalman & 63.7 & 67.1 & \underline{60.2} & 67.4 & 62.3 & 63.6\\
            & E-LPT~\cite{sirohi2021efficientlps} & \underline{66.4} & \underline{70.4} & - & \underline{69.5} & \underline{67.5} & \underline{67.9}\\
            & {\bf Ours} & \textbf{73.2} & \textbf{74.9} & \textbf{66.6} & \textbf{80.4} & \textbf{72.0} & \textbf{76.0}\\
        \bottomrule
        \end{tabular}
    }
    \label{tab:nuscenes_panotrack}
\end{table}

\subsection{Benchmark results} 
\label{sec:bench}
This section compares our method to published state-of-the-art methods on standard benchmarks for 3D and 4D lidar panoptic segmentation datasets~\cite{Behley19ICCV,fong21arxiv}. %

\PAR{Lidar panoptic segmentation.} We report the results for panoptic segmentation in Tab.~\ref{tab:nuscenes_panoseg} (Panoptic nuScenes) and Tab.~\ref{tab:semkitti_panoseg} (SemanticKITTI). We utilize temporal supervised (MAX) for obtaining object targets for the benchmark submission. We focus this discussion on the test set results and show results on the validation set for completeness. 
On the nuScenes dataset, \methodabr is the second-best method with $76.1$ $PQ$. 
Note that we only utilize standard convolution layers as opposed to proprietary Transformer-based Panoptic-PHNet~\cite{li2022panoptic}, so there is potential to replace our lightweight components with stronger transformer-based counterparts to achieve better performance. The end-to-end latency of over system is $169.9$ms. Moreover, Panoptic-PHNet~\cite{li2022panoptic} could not be easily extended for sequence-level scene understanding (ie, tracking), while as we will show next, \methodabr can achieve competitive performance on tracking by simply appending a greedy association module. \methodabr outperforms other approaches by a large margin ($+13.5$ $PQ$). On Semantic-KITTI, \methodabr is a close-second obtaining $61.0$ $PQ$, with state-of-the-art being $61.5$ $PQ$. This highlights that \methodabr generalizes well across different datasets. \methodabr also performs favorably against recent, query-based network~\cite{marcuzzi2023ral}, that extends state-of-the-art image-based approach, Mask2Former~\cite{cheng2022masked} to the lidar domain.

\PAR{Lidar panoptic tracking.} We report the results for 4D lidar panoptic tracking on SemanticKITTI and nuScenes datasets in Tab.~\ref{tab:nuscenes_panotrack}. Being a top-down method, \methodabr can easily extend to 4D panoptic segmentation through the greedy association of predicted velocity offsets. On nuScenes, \methodabr obtains $73.2$ LSTQ and $74.9$ PAT on the test set, establishing new state-of-the-art on this benchmark. \methodabr improves by $+6.8$ LSTQ and $+4.5$ PAT points over second-best approach, \textit{Efficient-LPT}~\cite{sirohi2021efficientlps}. On Semantic-KITTI, \methodabr obtains competitive results ($60.3$ LSTQ) with a simple greedy approach. 
These results affirm that \methodabr is a versatile approach that performs consistently across different benchmarks, 3D and 4D panoptic segmentation on multiple datasets. We refer the reader to the accompanying video for qualitative results.

\section{Conclusions}

This paper presents a \textit{top-down} approach to lidar panoptic segmentation and tracking using only \textit{modal} annotations. Our unified network jointly detects objects as modal points and classifies voxels to obtain per-point panoptic segmentation predictions. Instances are associated across 4D spatio-temporal data using learned modal velocity offsets to obtain panoptic tracking predictions. Our method establishes a new state-of-the-art on Panoptic nuScenes 4D panoptic segmentation benchmark. 
We hope that this work %
will inspire future developments in recognition-centric methods for lidar panoptic segmentation and tracking. 

\footnotesize{\PAR{Acknowledgments.} This project was funded by, in parts, by Sofja Kovalevskaja Award of the Humboldt Foundation.}

\bibliographystyle{IEEEtran}
\bibliography{IEEEabrv,egbib}

\end{document}

\title{Supplementary Material}%

\maketitle
\section{Inference Pseudocode}

In Alg.~\ref{algo:panoptic}, we present the pseudocode for inference using \methodabr to generate panoptic segmentation predictions (i.e. per-point semantic labels $A$ and instance ids $B$). 
The pseudocode assumes access to the following inputs:
\begin{itemize}
    \item the modal detection outputs $D$ (a list of detected modal centers including their predicted geometric coordinates $\hat{u}$, confidence $\hat{\sigma}$, class $\hat{k}$ and radius $\hat{r}$), 
    \item per-point semantic predictions $O^{point}$, 
    \item per-point features $F^{point}$, 
    \item per-point or per-center interpolated bird-eye-view features $F^{bev}$. 
\end{itemize}
We denote all points in raw point cloud as $P(x,y,z)$.
We first sort all the predicted centers in increasing order of confidence (line 3). Next, we generate the predictions for all \textit{thing} points (lines 5-12). We construct a RoI around each predicted center (line 6), ignoring points whose predicted semantic class label differs from the center class label (line 7). %
We then assign a unique instance id for those remaining points with membership confidence greater than $0.5$ (line 10). For all the \textit{stuff} points, we simply set the semantic predictions as those obtained from the semantic segmentation branch (line 13-15). 

\begin{algorithm}[ht]
\caption{Pseudocode for generating panoptic segmentation outputs}
\label{algo:panoptic}
\begin{algorithmic}[1]
\State \textbf{Input:} Modal Detections $D = [\{center: \hat{u}, conf.: \hat{\sigma}, class: \hat{k}, extent: \hat{r}\}, \dots]$, per-point semantic predictions $O^{point}$, per-point features $F^{point}$, per-point BEV features $F^{bev}$ for points in $P(x,y,z)$ and centers in $D$%
\State \textbf{Output:} Semantic Labels $A$, Instance Ids $B$
\vspace{1.5mm}

\State Sort predicted centers in $D = [\{\hat{u},\hat{\sigma}, \hat{k}, \hat{r}\}, \dots]$ in decreasing order of confidence $[\hat{\sigma}]$
\State Set $id = 0$  \Comment{instance id}
\For{each $d = [\hat{u}_d, \hat{\sigma}_d, \hat{k}_d, \hat{r}_d]$ in $D$} %
\State Construct $\mathcal{S} = \{p_i \in P(x,y,z) | \quad |p_i - \hat{u}_d| < \hat{r}_d\}$ \Comment{points within RoI}
\State Set $\mathcal{S} = \mathcal{S} \setminus \{p_i \in P(x,y,z) | \quad O^{point}_{i} == k_d\}$ \Comment{ignores points with different sem labels with center}
\State Construct $f^{point}_i = [p_i; F^{point}_i; F^{bev}_i; O^{point}_i] \quad \forall p_i \in \mathcal{S}$ \Comment{concat features for points}
\State Construct $f^{center} = [\hat{u}_d; F^{bev}; \hat{k}_d]$ \Comment{concat features for center}
\State Set $(A(p_i), B(p_i)) = (\hat{k}_d, id) \quad \forall p_i \in \mathcal{S} \quad st.~\text{\memfunc}(f^{point}_i, f^{center}) > 0.5$ %
\State $id = id + 1$
\EndFor
\For{each unassigned point $p_i \in P(x,y,z)$}
\State Set $(A(p_i), B(p_i) = (O^{point}(p_i), 0)$ \Comment{assign labels to stuff points}
\EndFor
\end{algorithmic}
\end{algorithm}

\section{Implementation Details}
Following CenterPoint~\cite{yin2021center}, we fix the voxel size as $0.075$, $0.075$ and $0.2$ for width, depth and height axis respectively. All points outside a range of $54m$ from the lidar sensor are ignored. This results in a 3D voxel grid of shape $1440\times1440\times32$. The input point cloud is downsampled by a factor of $8$ through $3$ downsampling residual blocks. The downsampling residual block comprises of two asymmetrical 3D convolutions~\cite{zhu2020cylindrical} with kernel size as $3\times1\times3$ and $1\times3\times3$, along with BatchNorm and ReLU layers. The 2D BEV feature extractor consists of two $3\times3$ convolutional layers, with BatchNorm and ReLU layers. These are then upsampled to the original resolution using upsampling blocks. The upsampling blocks apply a deconvolution operation with kernel size as $3\times3\times3$ on features from the skip connection, followed by asymmetrical residual block as in the encoder. We accumulate past $10$ scans as input to \methodabr. A one-cycle learning rate scheduler is used along with Adam optimizer for a total of 20 epochs, with the maximum learning rate set as $1e^{-3}$. %

\section{Class-wise Results}
We report detailed class-wise PQ results on NuScenes validation set in Tab.~\ref{tab:nuscenes_classwise}, and on SemanticKITTI test set in Tab.~\ref{tab:semantic_classwise}. We only present class-wise results for baselines that also provide such fine-grained result tables. 

On nuScenes, we observe that \methodabr significantly improves the performance on almost all \textit{things} classes, with the overall improvement of $+5.3$ in terms of $PQ^{Th}$ ($PQ$ for \textit{thing} classes) as compared to Panoptic-PHNet~\cite{li2022panoptic}. For SemanticKITTI, we observe that \methodabr improves over existing methods on stuff classes. 

\begin{table*}[ht]
    \centering
    \setlength{\tabcolsep}{5pt}
    \resizebox{\linewidth}{!}{%
        \begin{tabular}{cl|cccccccccccccccc|ccc}
        \toprule           
        & Method & \rotatebox[origin=c]{90}{barrier} & \rotatebox[origin=c]{90}{bicycle} & \rotatebox[origin=c]{90}{bus} & \rotatebox[origin=c]{90}{car} & \rotatebox[origin=c]{90}{constr-vehicle} & \rotatebox[origin=c]{90}{motorcycle} & \rotatebox[origin=c]{90}{pedestrian} & \rotatebox[origin=c]{90}{traffic-cone} & \rotatebox[origin=c]{90}{trailer} & \rotatebox[origin=c]{90}{truck} & 
         \rotatebox[origin=c]{90}{driveable} & \rotatebox[origin=c]{90}{other flat} & \rotatebox[origin=c]{90}{sidewalk} & \rotatebox[origin=c]{90}{terrain} & \rotatebox[origin=c]{90}{man-made} & \rotatebox[origin=c]{90}{vegetation} & $PQ^{Th}$ & $PQ^{St}$ &
        $PQ$\\
            \midrule
            & R.Net~\cite{Milioto19IROS} + Mask R-CNN~\cite{He17ICCV} &   40.3& 25.7 &51.7 &62.5 &14.6 &48.3 &38.8 &41.8 &32.7 &43.0 &77.1 &41.5 &59.2 &42.1 &58.9 &67.9& 39.9 & 57.8 & 46.6 \\ 
            & PanopticTrackNet~\cite{hurtado20arxiv} & 47.1 & 32.9& 57.9 &66.3 &22.8& 51.1 &42.8 &46.8 &38.9 & 51.0 &81.5 &42.3 &61.8 &45.1 &60.9 &70.9 & 45.8 & 60.4 & 51.4   \\ 
            & KPC~\cite{Thomas19ICCV} + Mask R-CNN~]\cite{He17ICCV}&
            46.7& 31.5& 56.8 &65.7& 21.9 &50.4 &41.6 &44.9 &37.6 &49.1 &83.5 &43.1 &63.5 &48.6& 73.9 &71.5 & 44.6 & 62.9 & 51.5
            \\
            & Efficient-LPS~\cite{sirohi2021efficientlps} & \underline{56.8} & 37.8 & 52.4 & 75.6 & 32.1 & 65.1 & 74.9 & 73.5 & 49.9 & 49.7 & 95.2 & 43.9 & 67.5 & \underline{52.8} & 81.8 & 82.4 & 56.8 & 70.6 & 62.0  \\
            & Panoptic-PHNet~\cite{li2022panoptic} & 53.5 & \underline{77.5} & \underline{75.4} & \underline{90.8} & \underline{48.6} & \underline{87.3} & \underline{91.0} & \underline{87.0} & \underline{56.5} & \underline{72.6} & \textbf{96.7} & \textbf{58.3} & \textbf{72.4} & \textbf{54.9} & \underline{88.7} & \underline{84.8} & \underline{74.0} & \textbf{75.9} & \underline{74.7}  \\
            \midrule
            & Ours (\methodabr) & \textbf{66.3} & \textbf{80.0} & \textbf{78.6} & \textbf{93.1} & \textbf{66.1} & \textbf{91.0} & \textbf{93.6} & \textbf{90.1} & \textbf{59.1} & \textbf{75.4} & \underline{96.0} & \underline{49.4} & \underline{68.5} & 52.2 & \textbf{88.9} & \textbf{86.1} & \textbf{79.3} & \underline{73.6} & \textbf{77.1}  \\
        \bottomrule
        \end{tabular}
    }
    \caption{NuScenes validation set class-wise PQ. Our method significantly outperform all prior works, especially for \textit{thing} classes. Best results are \textbf{bolded}, while second best are \underline{underlined}.}
    \label{tab:nuscenes_classwise}
\end{table*}

\begin{table*}[ht]
    \centering
    \setlength{\tabcolsep}{5pt}
    \resizebox{\linewidth}{!}{%
        \begin{tabular}{cl|ccccccccccccccccccc|ccc}
        \toprule
        & Method & \rotatebox[origin=c]{90}{car} &
        \rotatebox[origin=c]{90}{truck} &
        \rotatebox[origin=c]{90}{bicycle} & \rotatebox[origin=c]{90}{motorcycle} &  \rotatebox[origin=c]{90}{other-vehicle} & \rotatebox[origin=c]{90}{person} & \rotatebox[origin=c]{90}{bicyclist} & \rotatebox[origin=c]{90}{motorcyclist}  & \rotatebox[origin=c]{90}{road} &
        \rotatebox[origin=c]{90}{sidewalk} &
        \rotatebox[origin=c]{90}{parking} &  \rotatebox[origin=c]{90}{other-ground} & \rotatebox[origin=c]{90}{building} &  \rotatebox[origin=c]{90}{vegetation} & \rotatebox[origin=c]{90}{trunk} & \rotatebox[origin=c]{90}{terrain} &
        \rotatebox[origin=c]{90}{fence} &
        \rotatebox[origin=c]{90}{pole}& \rotatebox[origin=c]{90}{traffic-sign} & $PQ^{Th}$ & $PQ^{St} $ & 
        $PQ$\\
            \midrule
            & R.Net~\cite{Milioto19IROS} + P.P.~\cite{Lang19CVPR} & 66.9 &6.7 &3.1 &16.2& 8.8 &14.6 &31.8 &13.5 &90.6 &63.2 &41.3& 6.7 &79.2 &71.2 &34.6& 37.4 &38.2 &32.8 &47.4& 20.2 & 49.3 & 37.1 \\
            & PanopticTrackNet~\cite{hurtado20arxiv} & 70.8 &14.4 &17.8 &20.9 &27.4 &34.2 &35.4 &7.9 &\textbf{91.2} & 66.1 &50.3 &10.5 &81.8 &75.9 &42.0 &44.3 &42.9 &33.4 &51.1 &28.6 & 53.6 & 43.1 \\
            & KPC~\cite{Thomas19ICCV} + P.P.~\cite{Lang19CVPR} & 72.5& 17.2 &9.2 &30.8 &19.6& 29.9 &59.4 &22.8 &84.6 &60.1 &34.1 &8.8 &80.7& 77.6 &53.9 &42.2 &49.0 &46.2& 46.8 & 32.7& 53.1 & 44.5  \\
            & KPC~\cite{Thomas19ICCV} + PV-RCNN~\cite{shi2020pv} & 84.5 &21.9& 9.9 &34.2& 25.6 &51.1 &67.9 &43.8 &84.9 &63.6 &37.1 &8.4 &83.7 &78.3 &57.5 &42.3 &51.1 &51.0 &57.4 & 43.2 & 55.9 & 50.2 \\
            & Panoster~\cite{gasperini2020panoster} &
            84.0 &18.5 &36.4 &44.7 &30.1 &61.1 &69.2 &51.1 &90.2 &62.5 &34.5 &6.1 &82.0& 77.7& 55.7& 41.2 &48.0 &48.9 &59.8 & 49.4 & 55.1 & 52.7 \\  
            & DS-Net~\cite{hong2021lidar} & \underline{91.2} & 28.8 & 45.4 & 47.2 & 34.6 & 63.6 & 71.1 & \underline{58.5} & 89.1 & 61.2 & 32.3& 4.0 & 83.2 & 79.6 & \underline{58.3} & 43.4 & 50.0 & 55.2 & 65.3 & 55.1 & 56.5 & 55.9\\
            & Efficient-LPS~\cite{sirohi2021efficientlps} & 85.7 & 30.3 & 37.3 & 47.7 & 43.2 & 70.1 & 66.0 & 44.7 & 91.1 & \textbf{71.1} & \textbf{55.3} & \underline{16.3} & \underline{87.9} & \underline{80.6} & 52.4 & \underline{47.1} & \underline{53.0} & 48.8 & 61.6 & 53.1 & \underline{60.5} & 57.4\\
            & GP-S3Net~\cite{razani2021gp} & 84.1 & \underline{36.7} & 41.8 & \textbf{77.0} & 42.1 & 73.0 & \textbf{81.1} & \textbf{84.2} & \underline{91.1} & 64.5 & 38.9 & 11.9 & 80.4 & 68.1 & 53.2 & 37.2 & 46.9 & \underline{55.7} & \textbf{72.4} &\textbf{65.0} & 56.4 & 60.0\\
            & Panoptic-PHNet~\cite{li2022panoptic} & \textbf{94.0} & \textbf{45.1} & \textbf{54.6} & \underline{62.4} & \textbf{51.2} & \underline{74.4} & \underline{76.3} & 52.0 & 89.9 & \underline{70.6} & 49.4 & 11.7 & 87.8 & 79.4 & 57.2 & 45.0 & 52.6 & 54.5 & 61.2 & \underline{63.8} & 59.9 & \textbf{61.5}\\
            \midrule
            & Ours (\methodabr) & 89.9 & 20.6 & \underline{49.8} & 56.2 & \underline{46.6} & \textbf{76.5} & 76.0 & 48.8 & 87.8 & 68.3 & \underline{51.8} & \textbf{18.2} & \textbf{88.0} & \textbf{82.4} & \textbf{63.8} & \textbf{47.8} & \textbf{58.0} & \textbf{60.0} & \underline{68.8} & 58.1 & \textbf{63.2} & \underline{61.0}   \\
        \bottomrule
        \end{tabular}
    }
    \caption{SemanticKITTI test set class-wise PQ. \methodabr performs really well on stuff classes, while being second best overall. Best results are \textbf{bolded}, while second best are \underline{underlined}.}
    \label{tab:semantic_classwise}
\end{table*}

\section{Visual Results}
\begin{figure*}[!ht]
\centering
\setlength{\fboxsep}{0.3pt}%
\centering
\includegraphics[width=0.8\linewidth]{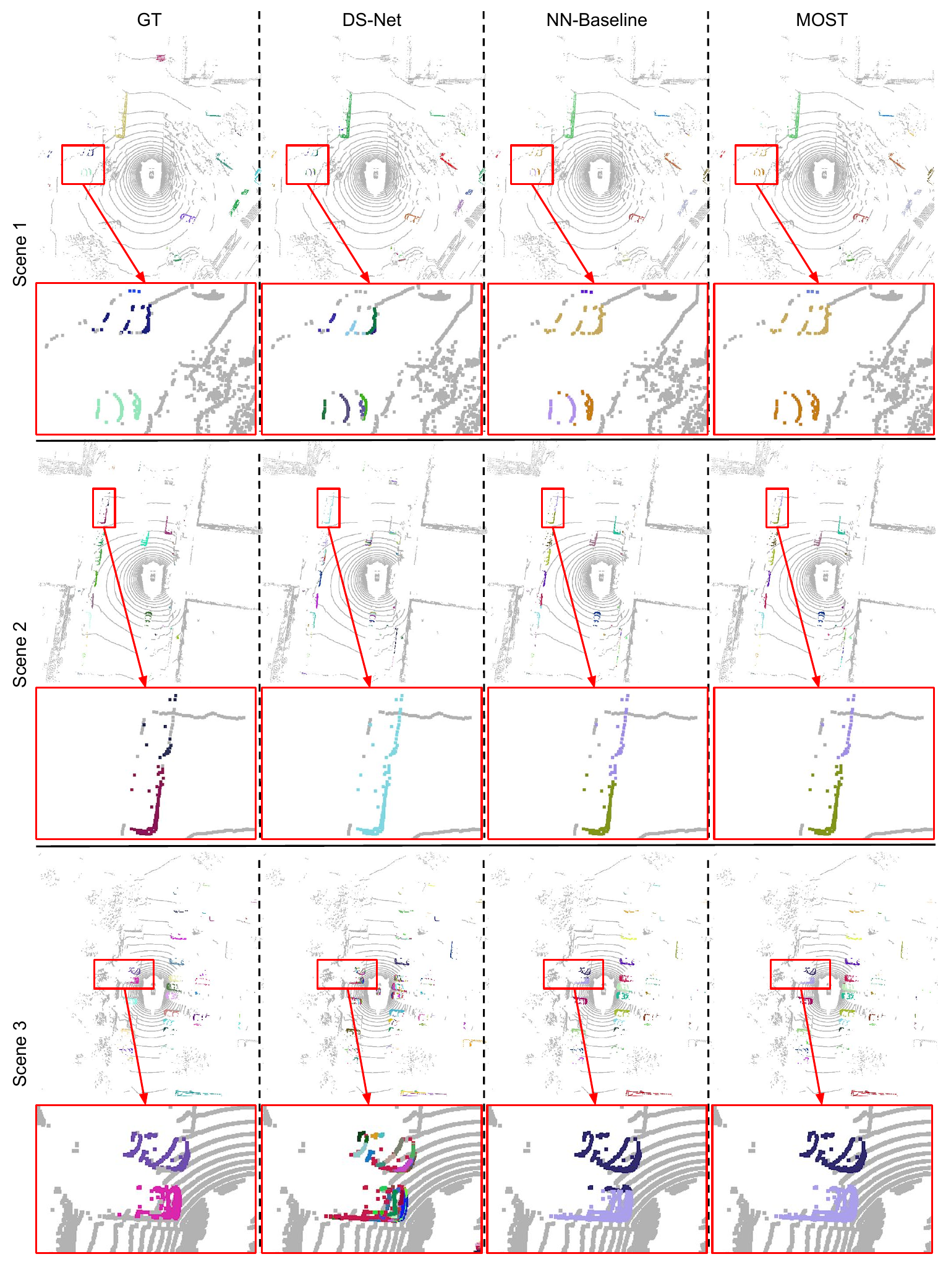}
\caption{Qualitative comparison of DS-Net~\cite{hong2021lidar}, our NN-Baseline and \methodabr. Only instance labels are visualized, and different colors represent different instances. We observe that DS-Net either fragments a single instance into multiple instances (Scene 1 and 3), or merge multiple instances into one (Scene 2). On the other hand, our NN-Baseline works well but can still fragment instances and fail to predict the boundary between instances, while \methodabr handles all these scenarios really well. } 
\label{fig:oneframe}
\end{figure*}

\begin{figure*}[!ht]
\centering
\setlength{\fboxsep}{0.3pt}%
\centering
\includegraphics[width=0.8\linewidth]{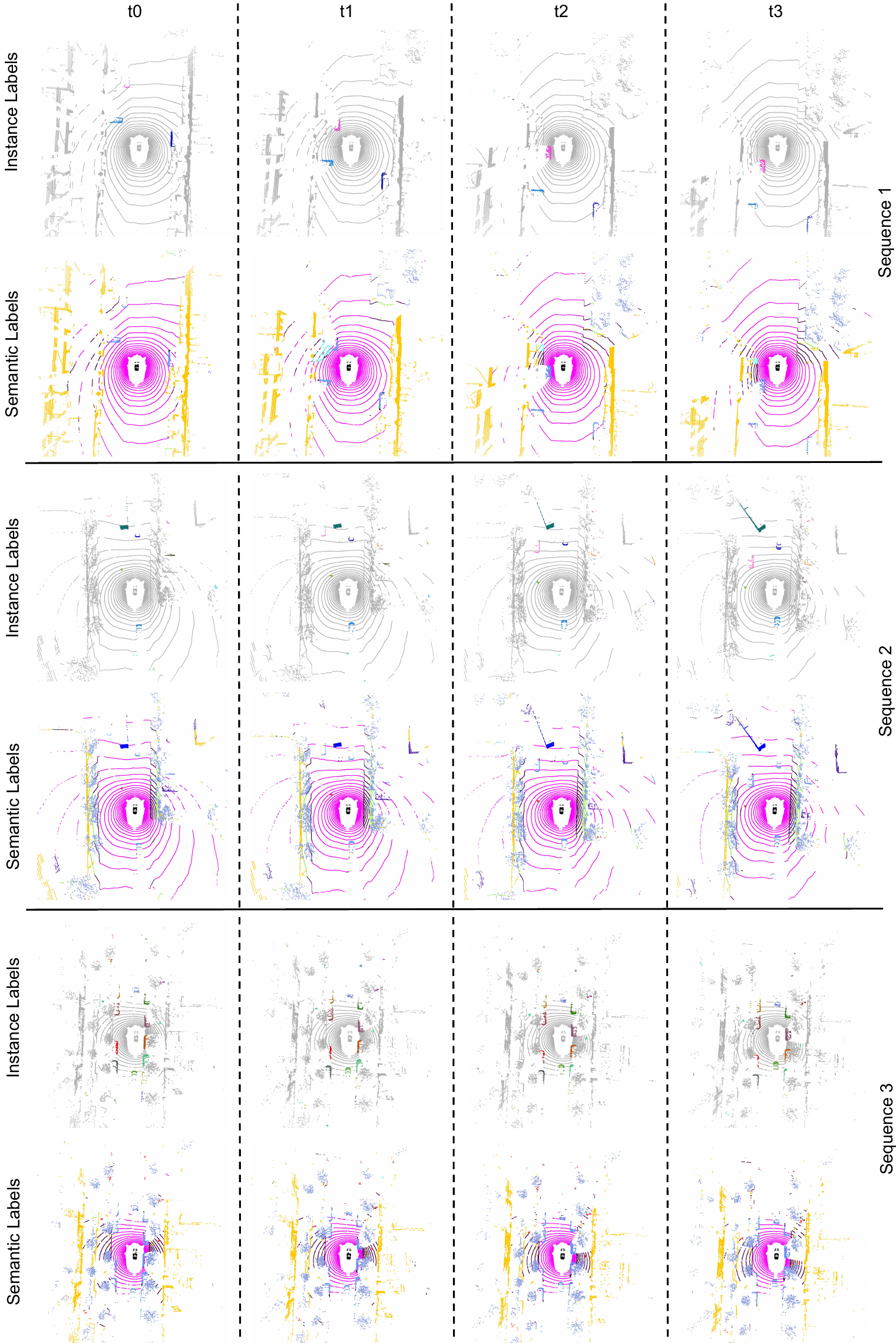}
\caption{Qualitative results of our 
\methodabr for 4D panoptic segmentation in successive frames of a sequence. We visualize the semantic and instance predictions for four successive frames of three different sequences. All the unique instance ids are mapped to a unique color across time. We observe that \methodabr can track objects really well for simple (Sequence 1 and 2) as well as more complex (Sequence 3) sequences.} 
\label{fig:successiveframes}
\end{figure*}

We qualitatively compare DS-Net~\cite{hong2021lidar}, our NN-baseline and \textit{MOST} in Fig.~\ref{fig:oneframe}. We only visualize \textit{instances} for clarity. The leftmost column visualizes ground truth instances, as shown with different colors. The middle columns represent the results for DS-Net and our NN-Baseline. The rightmost column visualizes our~\methodabr approach. Additionally, we visualize zoomed in versions of the scene for a detailed comparison of the approaches. We observe that DS-Net either fragments a single instance into multiple instances (Scene 1 \& 3), or merge multiple instances into one (Scene 2). On the other hand, our NN-Baseline works well but can still fragment instances (Scene 1) or lead to imprecise predictions (Scene 3). We observe that \methodabr handles all these scenarios really well leading to point-precise predictions.

We show qualitative results of \textit{MOST} for 4D panoptic segmentation in Fig.~\ref{fig:successiveframes}. We visualize the semantic and instance predictions for four successive frames of three different sequences. To visualize 4D predictions, all the unique instance ids are mapped to a unique color across time. We observe that \methodabr can track objects really well for simple (Sequence 1 \& 2) as well as more complex (Sequence 3) sequences. We refer the reviewers to the included video sequence generated using 4D panoptic segmentation predictions. We observe that \methodabr can track objects really well, except in a few cases where we observe identity switches.

\clearpage

\bibliographystyle{IEEEtran}
\bibliography{IEEEabrv,egbib}